\documentclass[10pt,twocolumn,letterpaper]{article}

\usepackage{amsmath,amsfonts,bm}

\def\eqref#1{eq.~\ref{#1}}

\def\1{\bm{1}}

\def\rvx{{\mathbf{x}}}

\def\Cbar{{\overline{C}}}

\def\rmM{{\mathbf{M}}}

\def\mM{{\bm{M}}}

\DeclareMathAlphabet{\mathsfit}{\encodingdefault}{\sfdefault}{m}{sl}
\SetMathAlphabet{\mathsfit}{bold}{\encodingdefault}{\sfdefault}{bx}{n}

\def\gX{{\mathcal{X}}}
\def\gY{{\mathcal{Y}}}

\def\sR{{\mathbb{R}}}

\newcommand{\E}{\mathbb{E}}

\DeclareMathOperator*{\argmax}{arg\,max}

\usepackage{iccv}
\usepackage{times}
\usepackage{epsfig}
\usepackage{graphicx}
\usepackage{amsmath}
\usepackage{amssymb}
\usepackage{varwidth}
\usepackage{IEEEtrantools}
\usepackage{booktabs}
\usepackage{diagbox}
\usepackage{floatrow}
\usepackage{bbding}
\usepackage{authblk}

\newcommand\blfootnote[1]{%
  \begingroup
  \renewcommand\thefootnote{}\footnote{#1}%
  \addtocounter{footnote}{-1}%
  \endgroup
}
\newtheorem{Theorem}{Theorem}

\allowdisplaybreaks

\iccvfinalcopy 

\title{Generative-Discriminative Complementary Learning}

\author[1]{\small Yanwu Xu \textsuperscript{*}}
\author[2]{\small Mingming Gong \textsuperscript{*}}
\author[1]{\small Junxiang Chen}
\author[4]{\small Tongliang Liu}
\author[3]{\small Kun Zhang}
\author[1]{\small Kayhan Batmanghelich\Envelope}
\affil[1]{\small Department of Biomedical Informatics, University of Pittsburgh, \{yanwuxu,juc91,kayhan\}@pitt.edu}
\affil[2]{\small School of Mathematics and Statistics, The University of Melbourne,  mingming.gong@unimelb.edu.au}
\affil[3]{\small Department of Philosophy, Carnegie Mellon University,  kunz1@cmu.edu}
\affil[4]{\small School of Computer Science, The University of Sydney, tongliang.liu@sydney.edu.au}

\begin{document}

\maketitle

\begin{abstract}

Majority of state-of-the-art deep learning methods are discriminative approaches, which model the conditional distribution of labels given inputs features. 
The success of such approaches heavily depends on high-quality labeled instances, which are not easy to obtain, especially as the number of candidate classes increases.
In this paper, we study the complementary learning problem.
Unlike ordinary labels, complementary labels are easy to obtain because an annotator only needs to provide a yes/no answer to a randomly chosen candidate class for each instance.  
We propose a generative-discriminative complementary learning method that estimates the ordinary labels by modeling both the conditional  (discriminative) and instance (generative) distributions.  
Our method, we call Complementary Conditional GAN ($CCGAN$), improves the accuracy of predicting ordinary labels and is able to generate high-quality instances in spite of weak supervision. In addition to the extensive empirical studies, 
we also theoretically show that our model can retrieve the true conditional distribution from the complementarily-labeled data.

\blfootnote{$^{\star}$Equal Contribution}
\end{abstract} 

\section{Introduction}
Deep supervised learning has achieved great success in various applications such as visual recognition \cite{krizhevsky2012imagenet,resnet} and natural language processing \cite{kim2014convolutional}. Despite the effectiveness of supervised classifiers, acquiring labeled data is often expensive and time-consuming. As a result, learning from weak supervision has been studied extensively in recent decades, including but not limited to semi-supervised learning~\cite{semi-gan}, multi-instance learning~\cite{zhou2012multi}, learning from side information~\cite{side-learning}, and learning from data with noisy labels~\cite{noisy-label}.

In this paper, we consider a recently proposed weakly-supervised classification scenario, \ie, learning from complementary labels \cite{lc,bc}. Unlike an ordinary label, a complementary label specifies a class that an input instance does {\it not} belong to. Given an instance from a class, it is laborious to choose the correct class label from many candidate classes, especially when the number of classes is relatively large or the annotator is not familiar with the characteristics of all candidate classes. However, it is less demanding and inexpensive to choose one of the incorrect class as a complementary label for an instance. 
For example, when an annotator is labelling an image containing an animal that she has never seen before, she can easily identify that this animal does not belong the usual animal classes he can see in daily life, such as, ``not dogs".
In medical field, a doctor may not be able to identify the exact disease type given symptoms. However, he/she can easily obtain complementary labels denoting some disease types a patient does not belong to.

Existing complementary learning methods modified the ordinary classification loss functions to enable learning from complementary labels. \cite{lc} proposed a method that provides a consistent estimate of the classifier from complementarily-labeled data where the loss function satisfies a particular symmetric condition. However, this method only allows classification loss functions with certain non-convex binary losses for one-versus-all and pairwise comparison. Later, \cite{bc} proposed to use the forward loss correction technique~\cite{patrini2017making} that learns the conditional, $P_{Y|X}$, from complementary labels, where $X$ denote the input features and $Y$ denote labels.  \cite{ac} derived an unbiased estimator of the true classification risk with arbitrary loss functions from complementarily-labeled data.

To clarify the differences between learning with ordinary and complimentary labels, we define the notion of ``effective sample size'', which is the number of instances with ordinary labels that carries the same amount of information as instances with complementary labels of a given size. 
Since the complementary labels are weak labels, they carry only partial information about the ordinary labels. 
Hence, the effective sample size $n_{l}$ for complementary learning
is much smaller than the given sample size $n$ (i.e., $n_{l} << n$). 
Current methods for learning with complementary labels need a relatively large training set to ensure low variance for predicting ordinary label.

Although $n_l$ is small under complementary learning settings, we can still use all samples with size $n$ to estimate the instance distribution $P_X$. However, current complementary methods focus on modeling conditional $P_{Y|X}$ and thus fail to account for information hidden in $P_X$, which is essential in complementary learning.

To improve the prediction performance, we propose a generative-discriminative complementary learning approach that learns both $P_{Y|X}$ and $P_{X|Y}$ in a unified framework. Our main contributions can be summarized as follows:
\begin{itemize}
    \item We propose a Complementary Conditional Generative Adversarial Net ($CCGAN$), which simultaneously learns $P_{Y|X}$ and $P_{X|Y}$ from complementary labels. Because the estimate of $P_{X|Y}$ benefits from $P_X$, it provides constraints on $P_{Y|X}$ and helps reduce its estimation variance.
    \item Theoretically, we show that our $CCGAN$ model is guaranteed to learn $P_{X|Y}$ from complementarily-labeled data.
    \item Empirically, we conduct comprehensive experiments on benchmark datasets, including MNIST, CIFAR10, CIFAR100, and VGG Face; demonstrating that our model gives accurate classification prediction and generates high-quality images.
\end{itemize}

\section{Related Works}

\paragraph{Generative Adversarial Nets}
Generative Adversarial Nets (GANs) are a class of implicit generative models learned by adversarial training~\cite{gans}. With the development of new network architectures~ (\eg,~\cite{brock2018large}) and stabilizing techniques (\eg,~\cite{sngan}), GANs generates high-quality images that are indistinguishable from real ones. Conditional GANs (CGANs)~\cite{cgans} extend the GAN models to generate images given specific labels, which can be used to model the class conditional $P_{X|Y}$ (\eg, AC-GAN~\cite{acgans}, Projection cGAN~\cite{p-gan}, and TAC-GAN \cite{gong2019twin}). However, training of CGANs requires ordinary labels for the images, which are not available under the complementary learning settings. To the best of our knowledge, our proposed $CCGAN$ is the first conditional GAN that is trained with complementary labels. The most related works to us are the robust conditional GAN approaches that aim to learn a conditional GAN from labels corrupted by random noise \cite{thekumparampil2018robustness,kaneko2018label}. However, our method generates better quality images and more accurate prediction, by utilizing complementary labels.

\paragraph{Semi-Supervised Learning}
Under semi-supervised learning settings, we are provided a relatively small number of labeled data and plenty of unlabeled data. The basic assumption for the semi-supervised methods is that the knowledge on $P_X$ gained from unlabeled data carries useful information for inferring $P_{Y|X}$. This principle has been implemented in various forms, such as co-training~\cite{blum1998combining}, generative modeling~\cite{odena2016semi,kumar2017semi}, { \it etc}.
Inspired by the commonalities between complementary learning and semi-supervised learning, \ie, more data are available to estimate $P_X$ than $P_{Y|X}$; we propose to make use of $P_X$ to help infer $P_{Y|X}$ in complementary learning. 
\section{Background}
In this section, we first introduce the concept of learning from so-called complementary labels. Then, we discuss a state-of-the-art discriminative  complementary learning approach,~\cite{bc}, which is the most relevant to our method.

\subsection{Problem Setup}
Let two random variables $X$ and $Y$ denote the features and the labels, respectively. The goal of 
discriminative learning
is to infer a decision function (classifier) from independent and identically distributed  training set $\{\rvx_i,y_i\}_{i=1}^n\subseteq\mathcal{X}\times\mathcal{Y}$ drawn from an unknown joint distribution $P_{XY}$, where 
$X \in \gX = \sR^d$ and $Y \in \gY=\{1,\ldots,K\}$.
The optimal function, $f^*$, can be learned by minimizing the expected risk $R(f)=\E_{(X,Y)\sim P_{XY}}\ell(f(X),Y)$, where $\E$ denotes the expectation and $\ell$ denotes a classification loss function. Because $P_{XY}$ is unknown, we usually approximate $R(f)$ using its empirical estimation $R_n(f)=\frac{1}{n}\sum_{i=1}^n\ell(f(\rvx_i),y_i)$.

In the complementary learning setting, for each sample $\rvx$, we are given only a complementary label $\bar{y}\in\mathcal{Y}\setminus{y}$ which specifies a class that $\rvx$ does {\it not} belong to. That is to say, our goal is to learn $f$ that minimizes the classification risk $R(f)$ from complementarily-labeled data $\{\rvx_i,\bar{y}_i\}_{i=1}^n\subseteq\mathcal{X}\times\mathcal{Y}$ drawn from an unknown distribution $P_{X\bar{Y}}$, where $\bar{Y}$ denote the random variable for complementary label. 
The ordinary loss function, $\ell(\cdot,\cdot)$, cannot be used since we do not have access to the ordinary labels ($y_i$'s). In the following, we explain how discriminative learning can be extended in such scenarios. 

\subsection{Discriminative Complementary Learning}
Existing Discriminative Complementary Learning ($DCL$) methods modified the ordinary classification loss function $\ell$ to the complementary classification loss $\bar{\ell}$ to provide a consistent estimation of $f$. Various loss functions have been considered in the literature, such as one-vs-all ramp/sigmoid loss  \cite{lc}, pair-comparison ramp/sigmoid loss  \cite{lc}, and cross-entropy loss  \cite{bc}. Here we briefly review a recent method that modifies the cross-entropy loss for deep learning with complementary labels \cite{bc}.
The general idea is to view the ordinary label $Y$, as a latent random variable.
Suppose the classifier has the form $f(X)=\argmax_{i\in[K]}g_i(X)$, where $g_i(X)$ is an estimation for $P(Y=i|X)$. The loss function for complementary labels is defined as $\bar{\ell}(f(X),\bar{Y})=\ell(\mM^\intercal \mathbf{g},\bar{Y})$, where $\mathbf{g}=(g_1(X),\ldots,g_K(X))^\intercal$ and $\mM$ is the transition matrix satisfying
\begin{eqnarray}\label{forming M}
 P(\bar{Y}=j|X)=\underset{i \not = j}{\sum} \underset{\mM_{ij}}{\underbrace{p(\bar{Y}=j|Y=i)}}P(Y=i|X).
 \label{Eq:relation}
\end{eqnarray}
\cite{lc,ac} assumed the uniform setting in which $\rmM$ takes 0 on diagonals and $\frac{1}{K-1}$ on non-diagonals. \cite{bc} relaxed this assumption by allowing other values on non-diagonals and proposed a method to estimate $\rmM$ from data.
It has been shown in \cite{bc} that the classifier $\bar{f}_n$ that minimizes the empirical estimation of $\bar{R}(f)$, i.e., 
\begin{equation}\label{Eq:bar_rnf}
\bar{R}_n(f)=\frac{1}{n}\sum_{i=1}^n\bar{\ell}(f(\rvx_i),\bar{y}_i),
\end{equation}
converges to the optimal classifier $f^*$ as $n\rightarrow\infty$.

\section{Proposed Method}

In this section, we will present the motivation and details of our generative-discriminative complementary learning method. First, we demonstrate why generative modeling is valuable for learning from complementary labels. Second, we present our Complementary Conditional GAN ($CCGAN$) model that is trained using complementarily-labeled data and provide theoretical guarantees. Finally, we discuss several practical factors that are crucial for reliably training our model.

\subsection{Motivation}
It is guaranteed that existing discriminative complementary learning approaches lead to optimal classifiers, given sufficiently large sample size. However, due to the uncertainty introduced by the complementary labels, the effective sample size is much smaller than the sample size $n$. If we have access to samples with ordinary labels $\{\rvx_i,y_i\}_{i=1}^n$, we can learn the classifier ${f}_n$ by minimizing $R_n(f)$. Since knowing the ordinary labels is equivalent to having all the $K-1$ complementary labels, we can also learn ${f}_n$ with ordinary labels by minimizing the empirical risk
\begin{equation}\label{Eq:bar_rnf'}
\bar{R}'_n(f)=\frac{1}{n(K-1)}\sum_{i=1}^n\sum_{k=1}^{K-1}\bar{\ell}(f(\rvx_i),\bar{\mathbf{y}}_{ik}),
\end{equation}
where $\bar{\mathbf{y}}_{ik}$ is the $k$-th complementary label for the $i$-th example. In practice, since we only have one complementary label for each instance, we are minimizing $\bar{R}_n(f)$ as shown in Eq.~(\ref{Eq:bar_rnf}), rather than $\bar{R}'_n(f)$. Note that $\bar{R}_n(f)$ approximates $\bar{R}'_n(f)$ by randomly picking up one complementary label for the $i$-th example, which implies that the effective sample size is roughly $n/(K-1)$. In other words, although we provide each instance a complementary label, the accuracy of the classifier learned by minimizing $\bar{R}_n$ is close to that of a classifier learned with $n/(K-1)$ examples with ordinary labels.

Because the effective sample size is usually much smaller than the actual sample size, complementary learning resembles semi-supervised learning, where only a small proportion of instances are associated with ordinary labels. In semi-supervised learning, $P_X$ can be estimated with more unlabeled samples compared to $P_{Y|X}$, which requires labels to estimate. Therefore, modeling $P_X$ is beneficial because it allows us to take advantage of unlabeled data. This justifies the motivation of introducing a generative term in complementary learning. A natural way to utilize $P_X$ is to model the class-conditional, $P_{X|Y}$. $P_X$ imposes a constraint on $P_{X|Y}$ indirectly since $P_X = \int P(X|Y=y)P(y)dy$. Therefore, a more accurate estimation of $P_X$ will improve the estimation of $P_{X|Y}$ and thus $P_{Y|X}$.

\subsection{Complementary Conditional GAN ($CCGAN$)}
 \begin{figure*}[!t]
 \vspace{-1.5cm}
 \begin{center}
   \includegraphics[width=15cm]{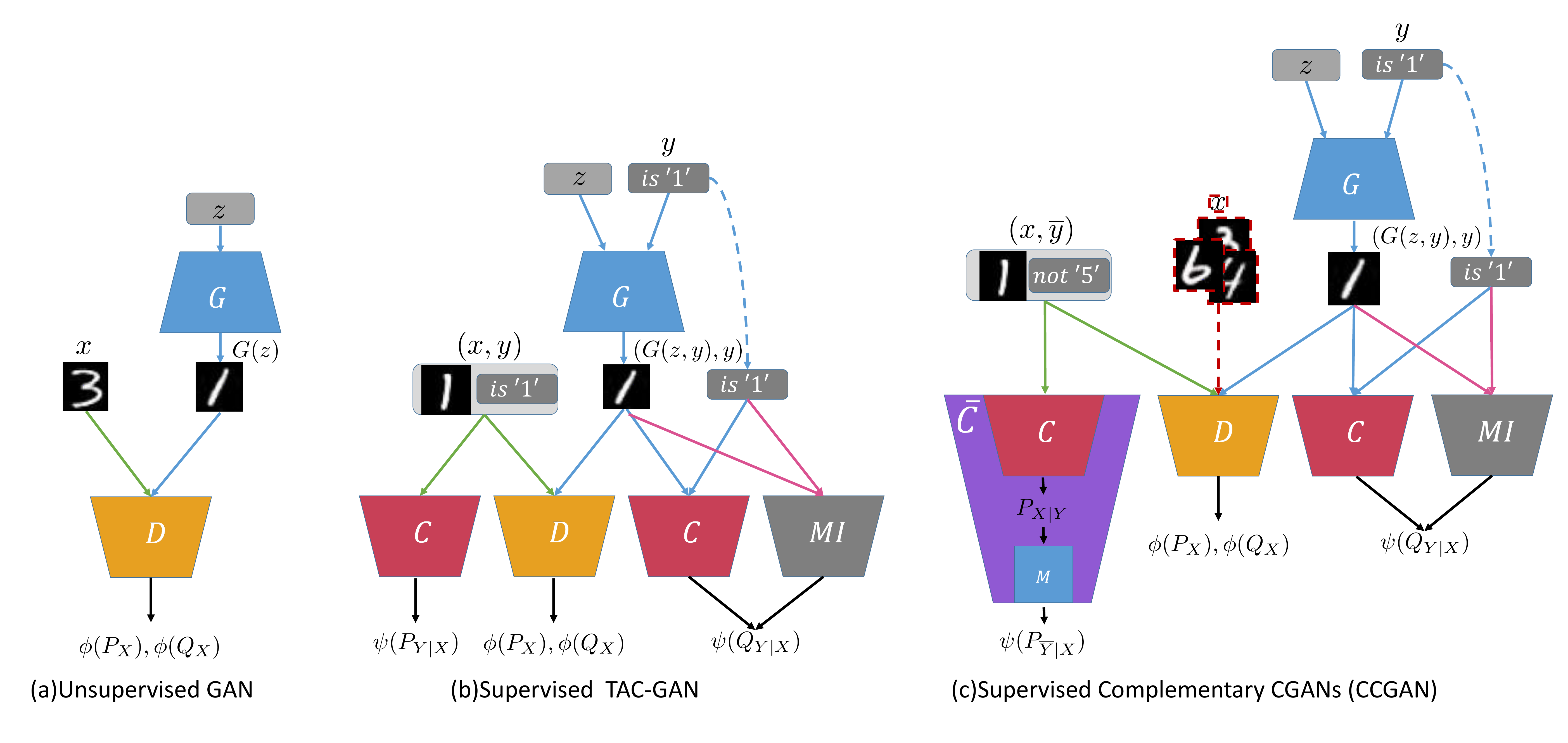}
 \end{center}
   \caption{\small GAN model with different supervision level. Figure (c) is our proposed method. The trapezoid blocks represent different networks, $G$ denotes Generator, $D$ denotes Discriminator, $C$ denotes true label classifier, $MI$ denotes mutual information learner and $\Cbar$ denotes complementary label classifier. In our model $CCGAN$, $C$ is regarded as latent output. As shown in (c),  $\Cbar$ can be decomposed into $C$ and transition matrix $\mM$ \cite{bc}, which can be referred to Eq.~(\ref{forming M}). }
 \label{model_structure}
 \end{figure*}
 
 Given the recent advances in generative modeling using  (conditional) GANs, we propose to use conditional GAN to model $P_{X|Y}$ in the paper. A conditional GAN learns a function $G(Y, Z)$ that generates samples from a conditional distribution $Q_{X|Y}$, neural network is used to parameterize the generator function, and $Z$ is a random samples drawn from a canonical distribution $P_Z$. To learn the parameters, we can minimize certain divergence between $Q_{X,Y}$ and $P_{XY}$ by solving the following optimization:
 \begin{flalign}\label{Eq:cgan}
 \min_G\max_D&\underset{(X,Y) \sim P_{XY}}{\E} [\phi( D(X,Y))]\nonumber\\&+\underset{Z\sim P_Z, Y\sim P_Y}{\E} [\phi(1- D(G(Z,Y),Y))],
 \end{flalign}
where $\phi$ is a function of choice and $D$ is the discriminator. 

However, the conditional GAN framework cannot be directly used for our purpose for the following two reasons: 1) the first term in Eq.~(\ref{Eq:cgan}) cannot be evaluated directly, because we do not have access to the ordinary labels. 
2) the conditional GAN only generates $X$'s and does not infer the ordinary labels.
A straightforward solution would be to generate $(\rvx, y)$ from the learned conditional GAN model and to train a separate classifier on the generated data. However, such two-step solution results in a sub-optimal performance.

To enable generative-discriminative complementary learning, we propose a complementary conditional GAN ($CCGAN$) by extending the TAC-GAN ~\cite{gong2019twin} framework to deal with complementarily-labeled data.  The model structures of GAN, TAC-GAN, and our $CCGAN$ are shown in Figure~\ref{model_structure}. TAC-GAN decomposes the joint distributions as $P_{XY}=P_{Y|X}P_X$ and $Q_{XY}=Q_{Y|X}Q_X$ and match the conditional distributions and marginal distributions separately. The marginals $P_X$ and $Q_X$ are matched using adversarial loss~\cite{gans}, and $P_{Y|X}$ and $Q_{Y|X}$ are matched by sharing a classifier with probabilistic outputs. 
However, $P_{Y|X}$ is not accessible in a complementary setting since the ordinary labels are not observed. Therefore, $P_{Y|X}$ and $Q_{Y|X}$ cannot be directly matched as in TAC-GAN.
Fortunately, we make use of the relation between $P_{Y|X}$ and $P_{\bar{Y}|X}$ ( Eq.~(\ref{Eq:relation}) ) and propose a new loss matching $P_{Y|X}$ and $Q_{Y|X}$ in a  complementary setting. Specifically, we learn our $CCGAN$ using the following objective
\begin{flalign}\label{Eq:main_obj}
  &\left.
      \begin{aligned}
      \min_{G,C} \max_{D,C^{mi}} &  \underset{X \sim P_X}{\E} \phi(D(X)) \\ &\hspace{1mm}+\underset{Z\sim P_Z, Y\sim P_Y}{\E} \phi(1-D(G(Z,Y))) \nonumber\\
    \end{aligned}
  \right\}
  \textcircled{ \small a }\\
  &\hspace{17.2mm}+ \underset{(X,\bar{Y})\sim P_{X\bar{Y}}}{\E} \ell(\bar{Y},\bar{C}(X) )
  \hspace{2mm}  \textcircled{\small b}\nonumber\\
  &\left.
  \begin{aligned}
  &\hspace{16.2mm}+\underset{Z\sim P_{Z},Y\sim P_{Y}}{\E} \ell(Y, C(G(Z,Y)))
  \hspace{2mm} \\
  &\hspace{16.2mm}+\underset{Z\sim P_{Z},Y\sim P_{Y}}{\E} \ell(Y, C^{mi}(G(Z,Y)))
  \end{aligned}
  \right\}
  \textcircled{\small c},
\end{flalign}
where $\ell$ is the cross-entropy loss, $C$ is a function modeled by a neural network with softmax layer as the final layer to produce class probability outputs, $\bar{C}(X)=\mM^\intercal C(X)$, and $C^{mi}$ is another function modeled by a neural network with class probability outputs.  From the objective function, we can see that our method naturally combines generative and discriminative components in a unified framework. Specifically, the component \textcircled{\small b} performs pure discriminative complementary learning on the complementarily-labeled data (only learns $C$), and the components \textcircled{ \small a } and \textcircled{ \small c } perform generative and discriminative learning simultaneously (learn both $G$ and $C$).

The three components in Eq.~(\ref{Eq:main_obj}) correspond to the following three divergences: 1) component \textcircled{ \small a } corresponds to Jensen-Shannon divergence between $P_X$ and $Q_X$, 2) component \textcircled{ \small b } represents KL divergence between $P_{\bar{Y}|X}$ and $Q'_{\bar{Y}|X}$, and 3) component \textcircled{ \small c } corresponds to KL divergence between $Q'_{Y|X}$ and $Q_{Y|X}$, where $Q'_{Y|X}$ is a conditional distribution of ordinary labels given features modeled by $C$ and  $Q'_{\bar{Y}|X}$ is a conditional distribution of complementary labels given features implied by $Q'_{Y|X}$ through the relation $Q'_{\bar{Y}|X}=\mM^\intercal Q'_{Y|X}$. The following theorem demonstrates that minimizing these three divergences in our objective can effectively reduce the divergence between $Q_{YX}$ and $P_{YX}$.
\begin{Theorem}\label{Theorem:theorem1}
Let $P_{YX}$ and $Q_{YX}$ denote the data distribution and the distribution implied by our model, respectively. Let $Q'_{Y|X}$ ($Q'_{\bar{Y}|X}$) denote the conditional distribution of ordinary (complementary) labels given features induced by the parametric model $C$. If $\mM$ is full rank, we have
\begin{flalign}
d_{TV}(P_{XY}, Q_{XY})
&\leq 2c_1\sqrt{d_{JS}(P_X, Q_X)}\nonumber\\&+c_2 \|\mM^{-1}\|_\infty \sqrt{d_{KL}({P}_{\bar{Y}|X},{Q}'_{\bar{Y}|X})}\nonumber\\&+c_2\sqrt{d_{KL}(Q_{Y|X},Q'_{Y|X})},
\end{flalign}
where $d_{TV}$ is the total variation distance, $d_{JS}$ is the Jensen-Shannon divergence, $d_{KL}$ is the KL divergence, and $c_1$ and $c_2$ are two constants.
\end{Theorem}
A proof of Theorem \ref{Theorem:theorem1} is provided in Section S1 of the supplementary file. An illustrative figure that shows the relations between the quantities in Theorem \ref{Theorem:theorem1} is also provided in Section S2 of the supplementary file.

\subsection{Practical Considerations} 
\noindent {\bf Estimating Prior $\mathbf{P_Y}$} \hspace{1mm} In our $CCGAN$ model, we need to sample the ordinary labels $y$ from the prior distribution $P_Y$, which needs to be estimated from complementary labels. Let $\bar{P}_{\bar{Y}} = [{P}_{\bar{Y}}  (\bar{Y}=1),\ldots,{P}_{\bar{Y}}(\bar{Y}=K)]^\intercal$ be the vector containing complementary label probabilities and  $\bar{P}_{Y}=[{P}_{Y}(Y=1),\ldots,{P}_{Y}(Y=K)]^\intercal$ be true label probabilities.
We estimate $\bar{P}_{Y}$ by solving the following optimization:
\begin{flalign}
\min_{\bar{P}_{Y}}\|\bar{P}_{\bar{Y}}-\mM^\intercal\bar{P}_{Y}\|^2,\nonumber\\ s.t. ~||\bar{P}_{Y}||_1=1~\text{and}~\bar{P}_Y[i]\geq 0.
\end{flalign}
This is a standard quadratic programming (QP) problem and can be easily solved using a QP solver.

\noindent{\bf Estimating $\mM$} \hspace{1mm} If the annotator is allowed to choose to assign either an ordinary label or a complementary label for each instance, the matrix $\mM$ will be unknown because of the possible non-uniform selection of the complementary labels. In \cite{bc}, the authors provided an anchor-based method to estimate $\mM$, we also follow the same technique. Please refer to \cite{bc} for more details.

\noindent{\bf Incorporating Unlabeled Data} \hspace{1mm}
In practice, we may have access to additional unlabeled data. We can readily incorporate such unlabeled data to improve the estimation of the first term in Eq.~(\ref{Eq:main_obj}), which further improves the learning of $G$ through the second term in Eq.~(\ref{Eq:main_obj}) and eventually improves the classification performance.

\section{Experiments}
To demonstrate the effectiveness of our method, we present a number of experiments examining different aspects of our method. After introducing the implementation details, we evaluate our methods on three datasets, including MNIST \cite{mnist}, CIFAR10, CIFAR100 \cite{cifar10}, and VGGFACE2~\cite{vggface2}. We compare classification accuracy of our $CCGAN$ with the state-of-the-art Discriminative  Learning ($DCL$) method \cite{bc} and show the capability of $CCGAN$ to generate good quality class-conditioned images from complementarily-labeled data. In addition, ablation studies based on MNIST are presented to give a more detailed analysis of our method. To be notified, we also have the Inception Score and Fréchet Inception Distance (FID) to measure the generative performance of our model, the result is shown in S3.

\subsection{Implementation Details}
\noindent{\bf  Label Generation} \hspace{1mm}
All the four datasets have ordinary class labels, which allows generating labels to evaluate our method. Following the procedure in \cite{lc}, the label for each image was obtained by randomly picking a candidate class and asking the labeler to answer ``yes" or ``no" questions. In this case, The candidate classes are uniformly assigned to each image, and therefore the transition matrix $\rmM$ satisfies $\mM_{i,j}=1/(K-1),i \not = j; \mM_{i,j}=0, i=j$. Also, data are usually biased, and the annotators also tend to hold biased choices based on their experience. Thus transition matrix $\mM$ could be biased \cite{bc}. For uniformed $\mM$ we assume $\mM$ is known. However, for biased $\mM$, we consider both cases when true $\mM$ is given, and $\mM$ needs to be estimated during training time. To be notified, when generating complementary data, we assume $\mM$ is known.

\noindent{\bf Training Details} \hspace{1mm}
We implemented our $CCGAN$ model in {\it Pytorch}. We trained our $CCGAN$ model in an end-to-end strategy, which means the classifier and GAN discriminator share the common bottom to neck conventional layers except for the final fully-connected softmax layer as well as mutual information learner. To train our $CCGAN$ model, we optimized the whole objective equation~\ref{Eq:main_obj} using Adam \cite{adam} with learning rate $2e-4$, $\beta_{1}=0.0$, $\beta_{2}=0.999$ for both $D$ and $G$ network, where we train 2 steps of $D$ and 1 step of $G$  in each iteration for 10,000 iteration in total. To train our baseline $DCL$ model, we apply the same training strategy as \cite{bc} for all dataset. For the additional VGGFACE2 dataset, we apply the same training settings as CIFAR100. We adopted data augmentation for all datasets except MNIST, where we first resized all images to $32 \times 32$ resolution, employed random croppings to change the image into $28\times28$ and then applied zero-padding to turn the image back with $32 \times 32$ resolution.

\subsection{MNIST}\label{MNIST experiment}
\vspace{-0.6cm}
\begin{center}
\begin{figure}[!t]
\vspace{-1.0cm}
\centerline{\includegraphics[width=7cm]{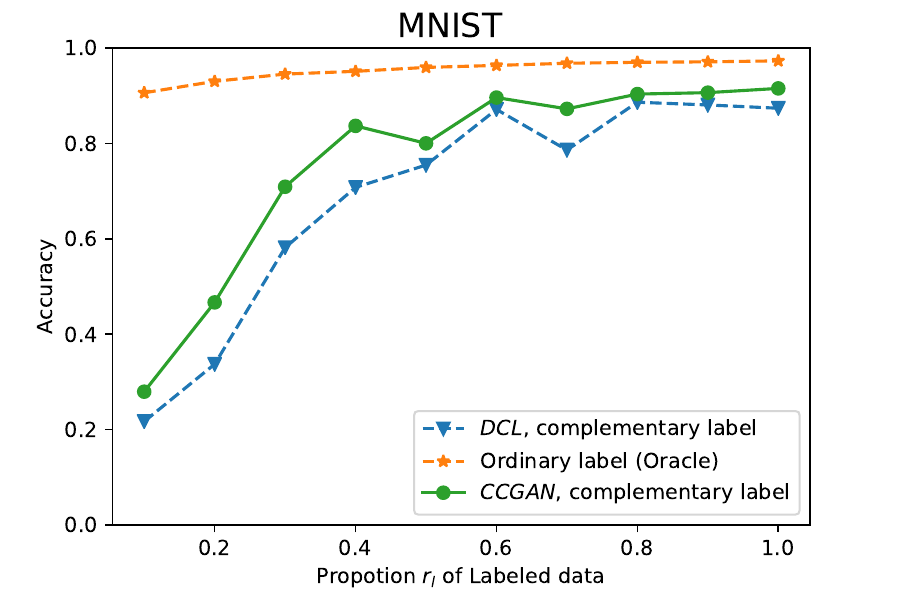}}
\centerline{(a)}
\centerline{\includegraphics[width=7cm]{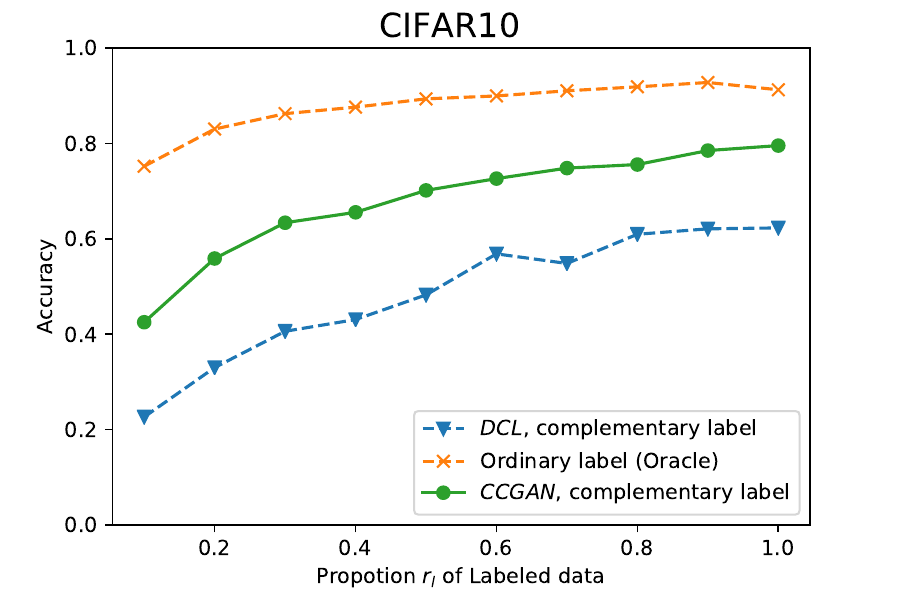}}
\centerline{(b)}
\caption{\small Test accuracy on (a) MNIST dataset and (b) CIFAR10 dataset. $x$ axis represents the proportion $r_l$ of labeled data in the training set $S$. In this figure we test $DCL$ and our proposed model $CCGAN$ for comparison. We also show the performance of ordinary classifier trained on ordinary labeled data as the Oracle. In this figure, transition matrix $mM$ is uniform and assumed to be known during training}\label{MNIST_CIFAR10_acc}
\vspace{-0.5cm}
\end{figure}
\end{center}

We first evaluate our model on MNIST, which is a handwritten digit recognition dataset that contains $60K$ training images and $10K$ testing images, with size $32\times 32$. We chose {\it Lenet-5} \cite{mnist} as the network structure for the $DCL$ method and the $C$ network in our $CCGAN$. We employed the DCGAN network \cite{dcgan} as the backbone of our $CCGAN$. Due to the simplicity of MNIST data, the accuracy of  learning is close to that of learning with ordinary labels if we use all $60K$ training samples. Therefore, we sample a subset of $6K$ images as our basic sampling set $S$ for training.

In the experiments, we evaluate all the methods under different sample sizes. Specifically, we randomly re-sampled subsets with $r_l\times 6K$ samples, where $r_l=0.1, 0.2, \ldots, 1$; and trained all the methods on these subsets. The classification accuracy was evaluated on the $10k$ test set. We report the results under the following three settings: 1) We only use samples with complementary labels, ignoring all ordinary labels, to train our model $CCGAN$ and baseline $DCL$. 2) We also train ordinary classifier such that all labeled data are provided with ordinary labels (Oracle). This classifier is trained with the strongest supervision possible,  representing the best achievable classification performance. The results are shown in Figure ~\ref{MNIST_CIFAR10_acc} (a).

\begin{figure*}
 \begin{center}
\vspace{-1.5cm}
   \includegraphics[width=14cm]{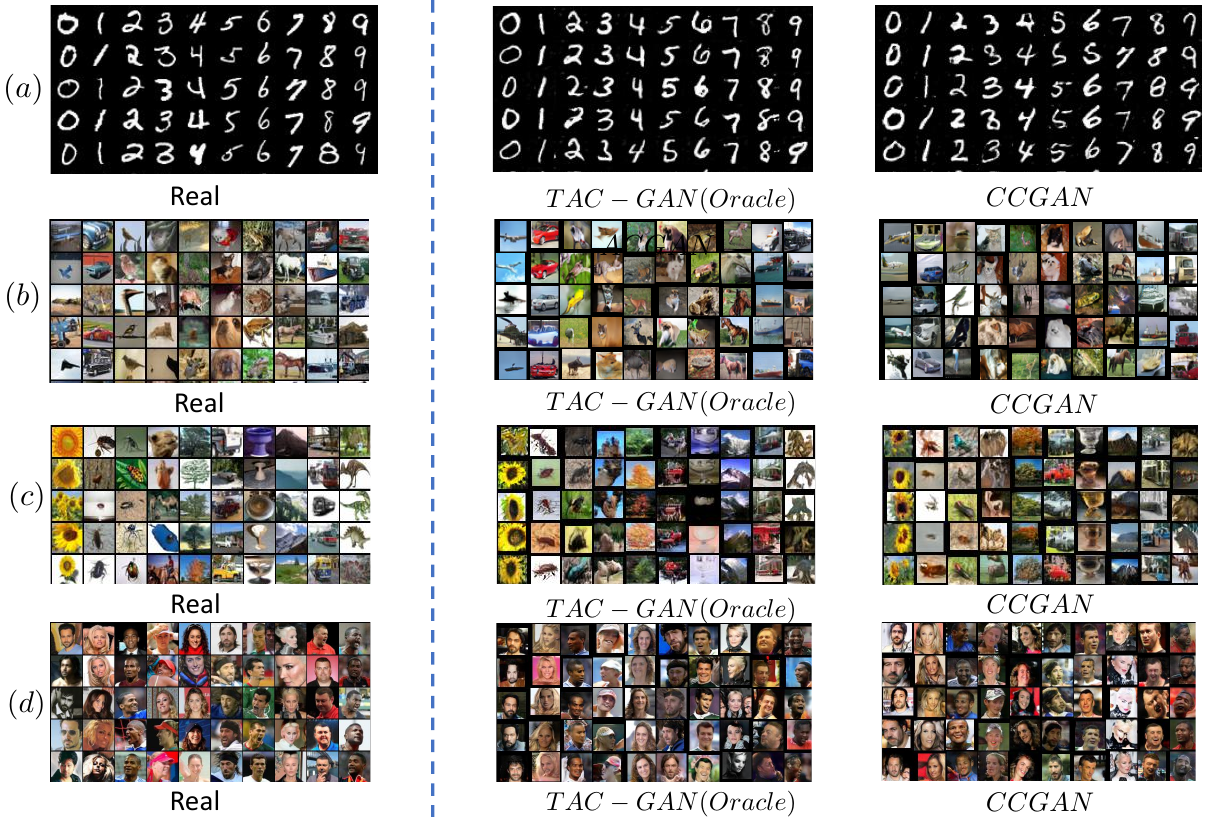}
 \end{center}
  \vspace{-0.35cm}
   \caption{\small Synthetic results for (a) MNIST (b) CIFAR10 (c) CIFAR100 and (d) VGGFACE100. The figures in the middle illustrate images generated by $TAC-GAN$~\cite{gong2019twin}, which is trained with ordinary labels. The figure on the right show images generated by $CCGAN$ model, which is trained with only complementary labels.}\label{all_sample}
   \vspace{-.3cm}
 \end{figure*}

It can be seen from the results that our $CCGAN$ method outperforms $DCL$ under different sample sizes, and the gap increases as the sample size reduces. our method outperforms $DCL$ by a large margin. 
The results demonstrate that generative-discriminative modeling is advantageous over discriminative modeling for complementary learning. Figure~\ref{all_sample} (a) shows the generated images from $TAC-GAN$ (Oracle) and our $CCGAN$. We can see that our $CCGAN$ generates high-quality digit images, suggesting that $CCGAN$ is able to learn $P_{X|Y}$ very well from complementarily-labeled data.

\subsection{CIFAR10}
We then evaluate our method on the CIFAR10 dataset, which consists of 10 classes of $32 \times 32$ RGB images, including $60K$ training samples and $10K$ test samples. We deploy {\it ResNet18} \cite{resnet} as the structure of the $C$ network in our model. Since training GANs on the CIFAR10 dataset is unstable, we utilize the latest conditional structure Big-GAN \cite{biggan} for our $CCGAN$ backbone. If without mention, the following dataset experiments apply the same settings. 

We evaluate all the methods following the same procedure used in the MNIST dataset. The results are shown in Figure ~\ref{MNIST_CIFAR10_acc} (b). Again our method consistently outperforms the $DCL$ method for different sample sizes. Figure~\ref{all_sample} (b) shows the generated images from $TAC-GAN$ (Oracle) and our $CCGAN$. It can be seen that our $CCGAN$ successfully learns the appearance of each class from complementary labels.

\begin{center}
\begin{table}
\addtolength{\tabcolsep}{-3pt}
\centering
\scalebox{0.85}{
 \begin{tabular}{l|c c c c c}
  \toprule
  \diagbox{Method}{$r_l$}  &0.2 &0.4&0.6 &0.8 & 1.0\\
  \midrule
 \multicolumn{6}{c}{VGGFACE100} \\
    \midrule
 Ordinary label (Oracle)&  0.673  & 0.804 & 0.870 & 0.891 & 0.917 \\
 \midrule
$DCL$ & 0.378 & 0.685 & 802 & 0.849 & 0.884 \\
$CCGAN$ &  \textbf{0.447}  & \textbf{0.728}  & \textbf{0.822} & \textbf{0.865}  & \textbf{0.896} \\
  \midrule
 \multicolumn{6}{c}{CIFAR100} \\
    \midrule
 Ordinary label (Oracle)&  0.439  & 0.804 & 0.870 & 0.891 & 0.917 \\
 \midrule
$DCL$ & 0.252 & 0.452 & 0.561 & 0.609 & 0.651 \\
$CCGAN$ &  \textbf{0.320}  & \textbf{0.520}  & \textbf{0.571} & \textbf{0.632}  & \textbf{0.660} \\
  \bottomrule
 \end{tabular}}
 \caption{\small This table shows the test accuracy on VGGFACE100 and CIFAR100 dataset. $r_l$ denotes the proportion of sampled labeled data for training from the training set $S$, best results are in bold}\label{all_table}
\end{table}
\end{center}

\vspace{-1cm}\subsection{CIFAR100 and VGGFACE100}
We finally evaluate our method on CIFAR100 and VGGFACE2 data, different from CIFAR10, CIFAR100 dataset contains 100 classes and each class has 500 images in average and 10.000 testing images of 100 classes in total. VGGCAE2 is a large-scale face recognition dataset. The face images have large variations in pose, age, illumination, ethnicity, and profession. The number of images for each person (class) varies from 87 to 843, with an average of 362 images for each person. We randomly sampled 100 classes and constructed a dataset for evaluation of our method. We selected $80\%$ data as the training set $S$ and the rest $20\%$ as the testing set. Since our $CCGAN$ model can only generate fixed-size images, we re-scaled all training images into $32\times 32$.

Because the number of classes is relatively large, the effective labeled sample size is approximately $n/99$, where $n$ is the total sample size. In case of limited supervision, neither $DCL$ nor our $CCGAN$ can converge. Thus, we applied the complementary label generation approach in \cite{bc}, which assumed only a small subset of candidate classes can be chosen as complementary labels. In specific, we randomly selected 10 candidate classes as the potential compelementary label each class, and assigned them with uniform probabilities.

We used the same evaluation procedures used in MNIST and CIFAR10. The classification accuracy is reported in Table~\ref{all_table}. It can be seen that our method outperforms $DCL$ by $5\%$ when the proportion of labeled data is smaller than 0.3 and is slightly better than $DCL$ when the proportion is larger than 0.5. Figure~\ref{all_sample} (c) shows the generated images from $TAC-GAN$ (Oracle) and our $CCGAN$. We can see that $CCGAN$ generates images that are visually similar to the real images for each person.

\subsection{Biased M training}
According to \cite{bc}, we also implement the biased transition matrix $\mM$ setting. During the training time, we test two settings: 1) we assume true $\mM$ used for generating data is known; 2) $\mM$ can not be acquired and needs to be estimated. For the unknown $\mM$, we follow the same settings as \cite{bc} and apply the same anchor method to estimate $\mM$. The other training settings are the same as above experiments. The result is shown in Table~\ref{biased_M}. 

\begin{center}
\begin{table}
\addtolength{\tabcolsep}{-3pt}
\centering
\scalebox{0.8}{
 \begin{tabular}{l|c c c | c c c}
  \toprule
  \diagbox{Method}{$r_l$}  &0.2 & 0.6 & 1.0& 0.2& 0.6& 1.0 \\
  \midrule
  & \multicolumn{3}{c}{True $M$} & \multicolumn{3}{c}{Esimated $M$}\\
 \midrule
 \multicolumn{7}{c}{MNIST} \\
 \midrule
$DCL$ & 0.675 & 0.866 & 880 & 0.563 & 0.787 & 0.894 \\
$CCGAN$ &  \textbf{0.839}  & \textbf{0.908}  & \textbf{0.918} & \textbf{0.773}  & \textbf{0.837} & \textbf{0.916} \\
  \midrule
 \multicolumn{7}{c}{CIFAR10} \\
 \midrule
$DCL$ & 0.413 & 0.658 & 0.724 & 0.282 & 0.624 & 0.713\\
$CCGAN$ &  \textbf{0.559}  & \textbf{0.767}  & \textbf{0.815} & \textbf{0.440}  & \textbf{0.740} & \textbf{0.757} \\
  \midrule
   \multicolumn{7}{c}{CIFAR100} \\
 \midrule
$DCL$ & 0.2814 & 0.582 & 663 & 0.176 & 0.381 & 0.574\\
$CCGAN$ &  \textbf{0.320}  & \textbf{0.621}  & 0.664 & \textbf{0.206}  & \textbf{0.445} & \textbf{0.589} \\
  \midrule
   \multicolumn{7}{c}{VGGFACE100} \\
 \midrule
$DCL$ & 0.461 & 0.769 & 0.863 & 0.161 & 0.660 & 0.836 \\
$CCGAN$ &  \textbf{0.533}  & \textbf{0.805}  & 0.866 & \textbf{0.174}  & \textbf{0.681} & \textbf{0.850} \\
  \midrule
  \bottomrule
 \end{tabular}}
 \caption{\small This table shows the test accuracy on MNIST, CIFAR10, CIFAR100, and VGGFACE100 when $M$ is biased, in this case, we implement our model when $M$ is known and estimated. $r_l$ denotes the proportion of sampled labeled data for training from the training set $S$.}\label{biased_M}
\end{table}
\end{center}

\subsection{Ablation Study}
Here we conduct ablation studies on MNIST to study the details and validate possible extensions of our approach.

\begin{figure}
\vspace{-1.5cm}
\centering
\begin{center}
  \centerline{\includegraphics[width=7cm]{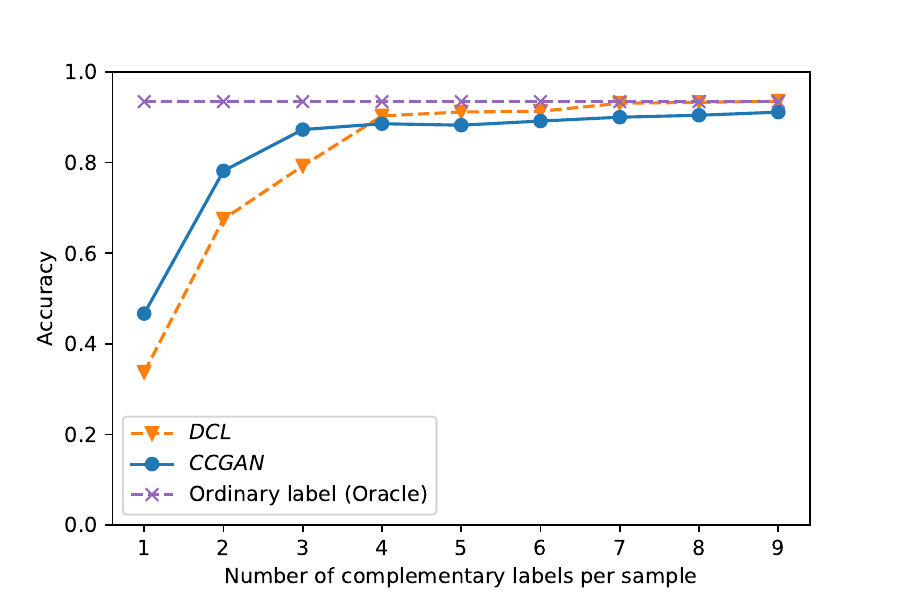}}
  \vspace{-0.5cm}
  \caption{\small Test accuracy. x axis denotes number of assigned complementary labels per image. In this figure we fix $r_l=0.2$ and . And we show the result of $DCL$ and our proposed model $CCGAN$, we also display the performance of ordinary classifier trained on ordinary labeled data as the Oracle.}\label{MNIST multi acc}
\end{center}
\end{figure}

\noindent{\bf Multiple  Labels} \hspace{1mm}
In this experiment, we give an intuitive strategy to verify the effectiveness of generative modeling for complementary learning. In ordinary supervised learning, discriminative models are usually preferred than generative models because estimating the high-dimensional $P_{X|Y}$ is difficult. To demonstrate the importance of generative modeling in complementary learning, we propose to assign multiple complementary labels to each image and observe how the performance changes with the number of complementary labels. The classification accuracy is shown in Figure~\ref{MNIST multi acc}. We can see that the accuracy of our $CCGAN$ and $DCL$ both increases with the number of complementary labels. When the number of complementary labels per image is large, $DCL$ performs better than our $CCGAN$ because the supervision information is sufficient. However, in practice, the number of complementary labels for each instance is typically small and is usually one. In this case, the advantage of generative modeling is obvious, as demonstrated by the superior performance of our $CCGAN$ compared to $DCL$.

\begin{figure}[t]
 \vspace{-1cm}
\centering
\begin{center}
  \centerline{\includegraphics[width=7cm]{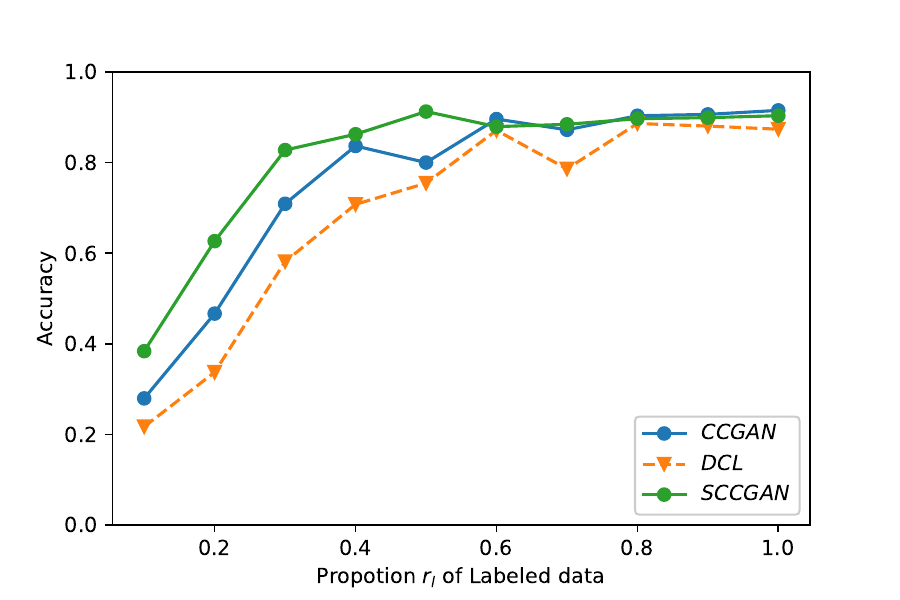}}
  \vspace{-0.5cm}
  \caption{\small Test accuracy. In this figure, we further compare our improved $SCCGAN$ model with our $CCGAN$ and $DCL$ model. x axis denotes the ratio $r_l$ of labeled data.}\label{MNISTS acc}
\end{center}
\vspace{-1cm}
\end{figure}

\noindent{\bf Semi-Supervised  Learning} \hspace{1mm}
In practice we might have easier access to unlabeled data which can be incorporated into to our model to perform semi-supervised complementary learning. On the MNIST dataset, we used the additional $90\%$ data as unlabeled data to improve the estimation of the first term in our objective Eq. (\ref{Eq:main_obj}). We denote the semi-supervised method as Semi-supervised complementary Conditional GAN($SCCGAN$). The classification accuracy w.r.t. different proportion of labeled data is shown in Figure~\ref{MNISTS acc}. We can see that $SCCGAN$ further improves the accuracy over $CCGAN$ due to the incorporation of unlabeled data.

\section{Conclusion}
We study the limitation of complementary learning as a weakly supervised learning problem, where the effective supervised information is much smaller compared to the sample size. To address this problem, we propose a generative-discriminative model to learn a better data distribution, as a strategy to boost the performance of the classifier. We build a conditional GAN model (CCGAN) which learns a generative model conditioned on ordinary class labels from complementary labeled data, and unify the generative and discriminative modeling in one framework. Our method shows superior classification performance on several datasets, including MNIST, CIFAR10, and CIFAR100 and VGGFACE100. Besides, our model generates high-quality synthetic images by utilizing complementary labeled data. In addition, we give a theoretical analysis that our model can converge to true conditional distribution learning from complementarily-labeled data.


{\small
\bibliographystyle{ieee}
\bibliography{generative_discriminative_complementary_learning_arxiv.bbl}
}

\onecolumn
\begin{center}
    \Large\textbf{Supplementary: ``Generative-Discriminative Complementary Learning"}\\
\end{center}

\:

\:

\:

\:

\:

\:

\maketitle


\setcounter{equation}{0}

\section*{S1. Proof of Theorem 1}
According to the triangle inequality of total variation (TV) distance, we have
\begin{flalign}
d_{TV}(P_{XY}, Q_{XY})&\leq d_{TV}(P_{XY}, P_{Y|X}Q_X) + d_{TV}(P_{Y|X}Q_X, Q_{XY}).\nonumber\\
\end{flalign}
Using the definition of TV distance, we have
\begin{flalign}
d_{TV}(P_{Y|X}P_X, P_{Y|X}Q_X)&=\frac{1}{2}\int |p_{Y|X}(y|x)p_X(x)-p_{Y|X}(y|x)q_X(x)|\mu(x,y)\nonumber\\
&\stackrel{(a)}{\leq} \frac{1}{2}\int |p_{Y|X}(y|x)|\mu(x,y)\int |p_X(x)-q_X(x)|\mu(x) \nonumber\\
&\leq c_1 d_{TV}(P_X, Q_X),
\end{flalign}
where $p$ and $q$ are densities, $\mu$ is a ($\sigma$-finite) measure, $c_1$ is a constant, and (a) follows from the H{\"o}lder inequality.

Similarly, we have 
\begin{equation}
d_{TV}(P_{Y|X}Q_X, Q_{Y|X}Q_X)\leq c_2 d_{TV}(P_{Y|X},Q_{Y|X}),
\end{equation}
where $c_2$ is a constant. Combining (1), (2), and (3), we have 
\begin{flalign}
d_{TV}(P_{XY}, Q_{XY})&\leq c_1 d_{TV}(P_X, Q_X)+c_2 d_{TV}(P_{Y|X},Q_{Y|X})\nonumber\\
&\leq c_1 d_{TV}(P_X, Q_X)+c_2 d_{TV}(P_{Y|X},Q'_{Y|X})+c_2 d_{TV}(Q'_{Y|X},Q_{Y|X}). 
\end{flalign}
Since we have no access to $P_{Y|X}$, by simply adapting the proof of Theorem 1 in \cite{thekumparampil2018robustness}, we bound $d_{TV}(P_{Y|X},Q'_{Y|X})$ using complementary conditional probabilities as
\begin{flalign}
d_{TV}(P_{Y|X},Q'_{Y|X})&=\underset{S_1,\ldots,S_K\subseteq \mathcal{X}}{\max}\sum_{y\in\mathcal{Y}}\{P(y|S_y)-Q'(y|S_y)\}\nonumber\\
&=\underset{S_1,\ldots,S_K\subseteq \mathcal{X}}{\max}\langle\mathbf{1},P(\cdot|\{S_y\}_{y\in\mathcal{Y}})-Q'(\cdot|\{S_y\}_{y\in\mathcal{Y}})\rangle\nonumber\\
&\stackrel{(a)}{=}\underset{S_1,\ldots,S_K\subseteq \mathcal{X}}{\max}\langle\mathbf{1},\mM^{-1}({P}(\cdot|\{S_{\bar{y}}\}_{\bar{y}\in\mathcal{Y}})-{Q}'(\cdot|\{S_{\bar{y}}\}_{\bar{y}\in\mathcal{Y}}))\rangle\nonumber\\
&\stackrel{(b)}{\leq}\|\mM^{-\intercal}\|_1\underset{S_1,\ldots,S_K\subseteq \mathcal{X}}{\max}\|{P}(\cdot|\{S_{\bar{y}}\}_{\bar{y}\in\mathcal{Y}})-{Q}'(\cdot|\{S_{\bar{y}}\}_{\bar{y}\in\mathcal{Y}}))\|_1\nonumber\\
&{=}\|\mM^{-1}\|_\infty d_{TV}({P}_{\bar{Y}|X},{Q}'_{\bar{Y}|X}),
\end{flalign}
where $P(\cdot|\{S_y\})=[P(Y=1|S_1),\cdots,P(Y=K|S_K)]^\intercal$, ${P}(\cdot|\{S_{\bar{y}}\})=[P(\bar{Y}=1|S_1),\cdots,P(\bar{Y}=K|S_K)]^\intercal$, (a) follows from ${P}(\cdot|\{S_{\bar{y}}\})=\mM P(\cdot|\{S_y\})$, and (b) follows from the fact that $\mathbf{1}^\intercal Ax\leq \|Ax\|_1\leq\|A\|_1\|x\|_1$. By combining (4) and (5), we have
\begin{flalign}
d_{TV}(P_{XY}, Q_{XY})
&\leq c_1 d_{TV}(P_X, Q_X)+c_2 \|\mM^{-1}\|_\infty d_{TV}({P}_{\bar{Y}|X},{Q}'_{\bar{Y}|X})\nonumber\\&~~+c_2 d_{TV}(Q_{Y|X},Q'_{Y|X})
\end{flalign}

According to the relations between total variation (TV), KL divergence ($d_{KL}$), and Jensen-Shannon divergence ($d_{JS}$), we can rewrite (6) as
\begin{flalign}
d_{TV}(P_{XY}, Q_{XY})
&\leq 2c_1\sqrt{d_{JS}(P_X, Q_X)}+c_2 \|\mM^{-1}\|_\infty \sqrt{d_{KL}({P}_{\bar{Y}|X},{Q}'_{\bar{Y}|X})}\nonumber\\&+~~c_2 \sqrt{d_{KL}(Q_{Y|X},Q'_{Y|X})},
\end{flalign}
which follows from the Pinsker's inequality. By replacing $2c_1$ in (7) with a new constant $c_1$ (using the same notation for simplicity), we can obtain the inequality in Theorem 1. From the theorem, we can see that if the complementary labels are highly-biased, it may cause $\mM$ to be rank-deficient. In this case, our algorithm may not minimize the distance between $P_{XY}$ and $Q_{XY}$ efficiently.

\section*{S2. Illustration of Our Objective Function (Eq. (5))}
\begin{figure}[h]
    \centering
    \includegraphics[width=0.6\textwidth]{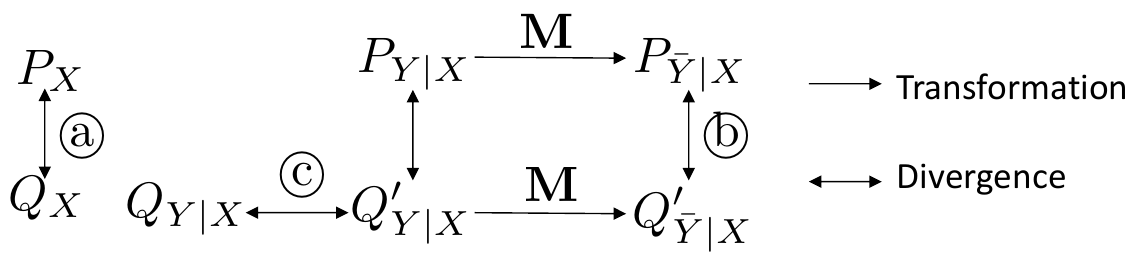}
    \caption{\small Illustration of the divergence terms that are minimized in Eq. (5). $P_{Y|X}$ ( $P_{\bar{Y}|X}$) is the conditional distribution of ordinal (complementary) label given features on the real data. $Q'_{Y|X}$ ( $Q'_{\bar{Y}|X}$) is the conditional distribution of ordinal (complementary) label produced by the classification network $C$ in Eq. (5). $Q_{Y|X}$ is the conditional distribution of ordinal label given features induced by our generator $G$. From the figure, we can see that minimizing $\textcircled{b}
$ leads to reduced divergence between $P_{Y|X}$ and $Q'_{Y|X}$. Therefore, the objective function minimizes the divergence between $P_{Y|X}$ and $Q_{Y|X}$ further because of $\textcircled{c}$. Combined with $\textcircled{a}
$, our objective minimizes divergence between $P_{XY}$ and $Q_{XY}$. }
    \label{fig:illus_objective}
\end{figure}
\section*{S3. Quality of synthetic data}

\begin{center}
\begin{table}[H]
\addtolength{\tabcolsep}{-3pt}
\scalebox{0.7}{
 \begin{tabular}{l|c c c c c}
  \toprule
  \diagbox{Method}{$r_l$}  &0.2 &0.4&0.6 &0.8 & 1.0\\
  \midrule
 \multicolumn{6}{c}{CIFAR10} \\
    \midrule
 Ordinary label, IS & 5.16 $\pm$ 0.066 & 5.99 $\pm$ 0.058 & 6.19 $\pm$ 0.070 & 6.27 $\pm$ 0.070 & 6.53 $\pm$ 0.082\\
$CCGAN$, IS & 5.28 $\pm$ 0.048 & 5.90 $\pm$ 0.065 & 6.27 $\pm$ 0.094 & 6.27 $\pm$ 0.067 & 6.48 $\pm$ 0.052\\
 Ordinary label, FID & 54.33 & 39.18 & 35.18 & 32.91 & 28.40\\
$CCGAN$, FID & 50.75 & 37.47 & 33.86 & 34.55 & 31.63\\
  \midrule
 \multicolumn{6}{c}{CIFAR100} \\
    \midrule
 Ordinary label, IS &   5.11 $\pm$ 0.038  & 6.80 $\pm$ 0.084  & 7.59 $\pm$ 0.154 & 7.94 $\pm$ 0.133  & 7.82 $\pm$ 0.09 \\
$CCGAN$, IS & 4.80 $\pm$ 0.042  & 6.36 $\pm$ 0.059  & 6.73 $\pm$ 0.095 & 7.17 $\pm$ 0.085  & 7.22 $\pm$ 0.115 \\
 Ordinary label, FID &   65.00  & 44.14  & 41.49 & 36.25  & 34.34 \\
$CCGAN$, FID & 79.13  & 44.01  & 43.63 & 36.21  & 34.63 \\
  \midrule
 \multicolumn{6}{c}{VGGFACE100} \\
    \midrule
 Ordinary label, IS &   19.18 $\pm$ 0.254  & 29.19 $\pm$ 0.235  & 48.99 $\pm$ 0.533 & 54.59 $\pm$ 0.390  & 67.77 $\pm$ 0.568 \\
$CCGAN$, IS & 16.49 $\pm$ 0.243  & 28.10 $\pm$ 0.368  & 45.82 $\pm$ 0.746 & 52.97 $\pm$ 0.470  & 62.30 $\pm$ 0.409 \\
 Ordinary label, FID &   100.48  & 66.00  & 42.98 & 38.07  & 26.26 \\
$CCGAN$, FID & 113.78  & 59.98  & 36.45 & 31.661  & 27.79 \\
  \bottomrule
 \end{tabular}}
 \hspace{0.7cm}
 \caption{ This table shows the Inception Score and FID socore on CIFAR10, CIFAR100 and VGGFACE100 dataset. $r_l$ denotes the proportion of sampled labeled data for training from the training set $S$. All these scores are under the uniformed $\mM$ setting.}\label{IS_FID}
\end{table}
\end{center}

\section*{S4. More Generated Images}
 \begin{figure*}[h]
\centering
\includegraphics[width=15cm]{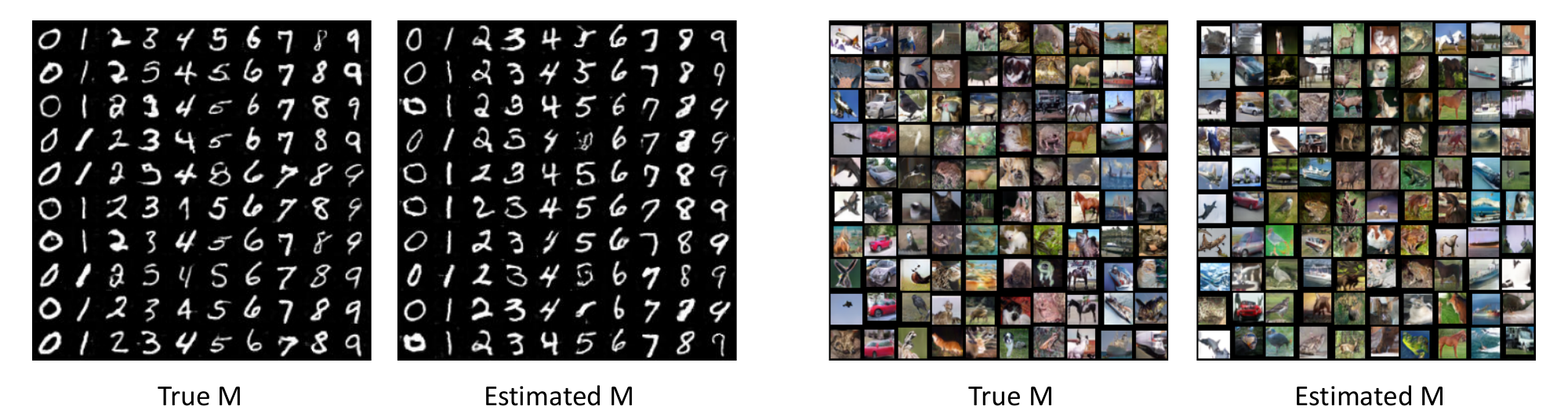}
  \caption{Synthetic results for MNIST and CIFAR10. We set $r_l=1$ here. It shows the generated data with true $\mM$ and esitimated $\mM$}
 \end{figure*}
 
  \begin{figure*}[h]
\centering
\includegraphics[width=15cm]{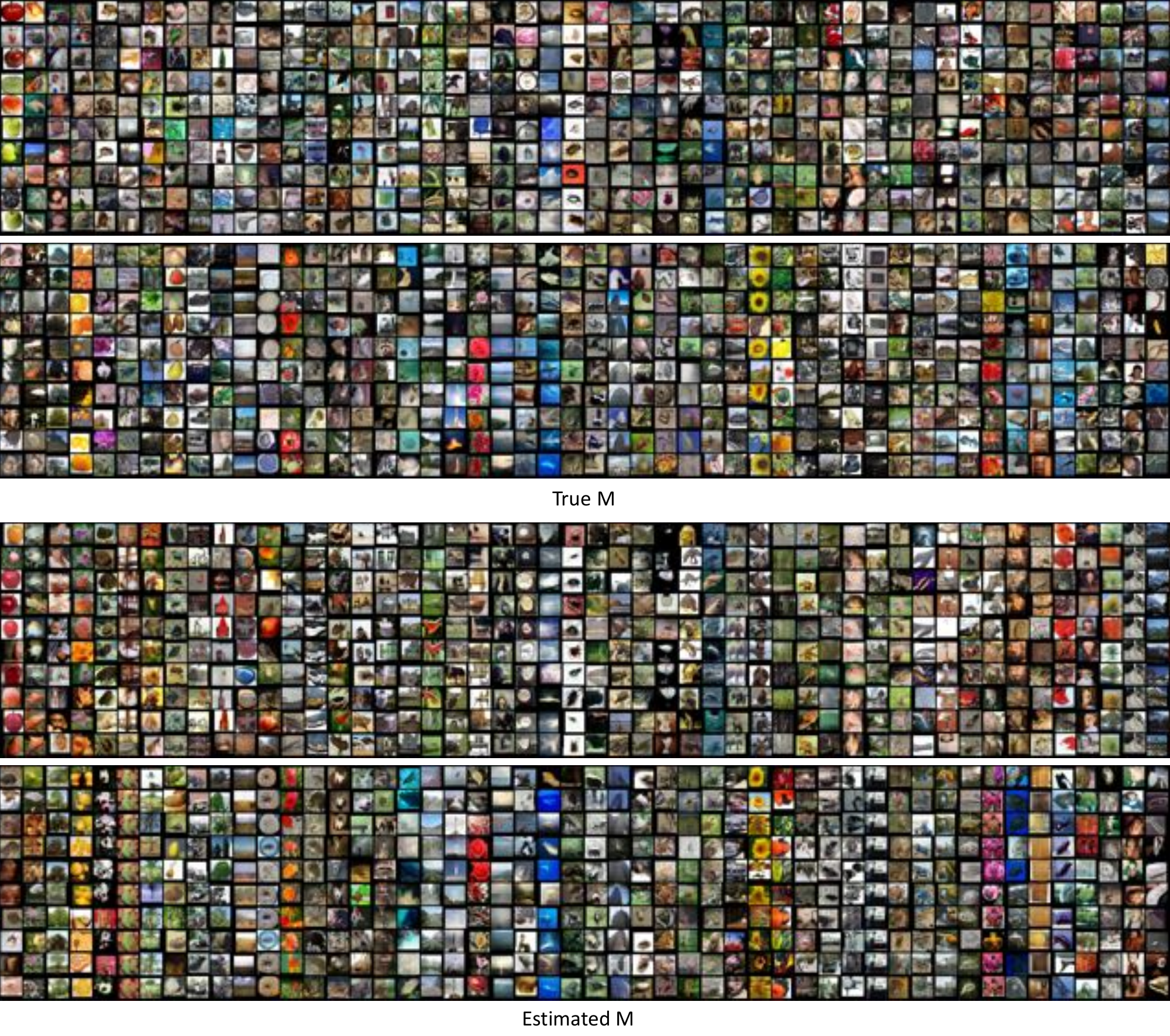}
  \caption{Synthetic results for CIFAR100. We set $r_l=1$ here. It shows the generated data with true $\mM$ and esitimated $\mM$}
 \end{figure*}
 
   \begin{figure*}[h]
\centering
\includegraphics[width=15cm]{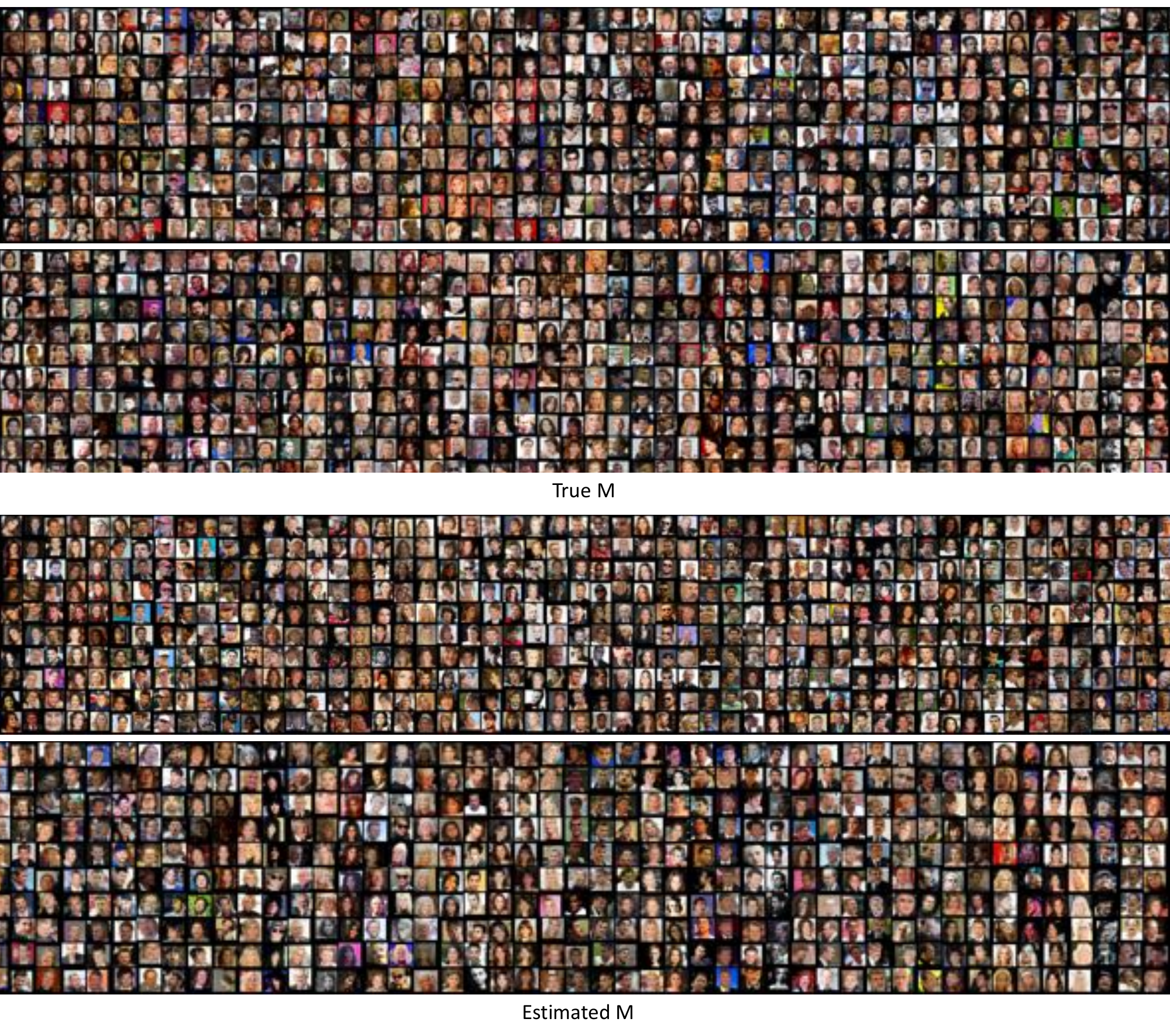}
  \caption{Synthetic results for MNIST and VGGFACE100. We set $r_l=1$ here. It shows the generated data with true $\mM$ and esitimated $\mM$}
 \end{figure*}




\end{document}